# Tempered Adversarial Networks

**Mehdi S. M. Sajjadi** [1 2]  **Giambattista Parascandolo** [1 2]  **Arash Mehrjou** [1]  **Bernhard Schölkopf** [1]

## Abstract

Generative adversarial networks (GANs) have been shown to produce realistic samples from high-dimensional distributions, but training them is considered hard. A possible explanation for training instabilities is the inherent imbalance between the networks: While the discriminator is trained directly on both real and fake samples, the generator only has control over the fake samples it produces since the real data distribution is fixed by the choice of a given dataset. We propose a simple modification that gives the generator control over the real samples which leads to a tempered learning process for both generator and discriminator. The real data distribution passes through a *lens* before being revealed to the discriminator, balancing the generator and discriminator by gradually revealing more detailed features necessary to produce high-quality results. The proposed module automatically adjusts the learning process to the current strength of the networks, yet is generic and easy to add to any GAN variant. In a number of experiments, we show that this can improve quality, stability and/or convergence speed across a range of different GAN architectures (DCGAN, LSGAN, WGAN-GP).

## 1. Introduction

Generative Adversarial Networks (GANs) have been introduced as the state of the art in generative models (Goodfellow et al., 2014). They have been shown to produce sharp and realistic images with fine details (Chen et al., 2016; Denton et al., 2015; Radford et al., 2016; Zhang et al., 2017). The basic setup of GANs is to train a parametric nonlinear function, the *generator G*, which maps samples from random noise drawn from a distribution $\mathcal{Z}$ into samples of a fake distribution $G(\mathcal{Z})$ which are close in terms of some measure to a real world empirical data distribution $\mathcal{X}$. To achieve this goal, a *discriminator D* is trained to provide feedback in the form of gradients for the generator. This feedback can be the confidence of a classifier discriminating between real and fake examples (Arjovsky et al., 2017; Goodfellow et al., 2014; Gulrajani et al., 2017; Mao et al., 2017) or an energy defined in terms of a reconstruction loss of an autoencoder (Berthelot et al., 2017; Zhao et al., 2017).

GANs are infamous for being difficult to train and sensitive to small changes in hyper-parameters (Goodfellow et al., 2016). A typical source of instability is the discriminator rapidly overpowering the generator which leads to problems such as vanishing gradients or mode collapse. In this case, $G(\mathcal{X})$ and $\mathcal{X}$ are too distant from each other and the discriminator learns to fully distinguish them (Arjovsky & Bottou, 2017). While several GAN variants have been introduced to address the problems encountered during training (Arjovsky et al., 2017; Berthelot et al., 2017; Gulrajani et al., 2017; Zhao et al., 2017), finding stable and more reliable training procedures for GANs is still an open research question (Lucic et al., 2017).

### 1.1. Our Contributions

In this work we propose a general and dynamic, yet simple to implement extension to GANs that encourages a smoother training procedure. We introduce a *lens* module $L$ which gives the generator control over the real data distribution $\mathcal{X}$ before it enters the discriminator. By adding the lens between the real data samples and the discriminator, we allow training to self-stabilize by automatically balancing a reconstruction loss with the current performance of the generator and discriminator. For instance, a lens could implement an image blurring operation which gradually gets reduced during training, thus only requiring the generation of good blurry images at the beginning, which gradually become sharper during training. While this analogy from optics motivates the term lens, in practice we learn the lens from data as explained below.

While the generator in a regular GAN chases a fixed distribution $\mathcal{X}$, the proposed lens moves the target distribution closer to the generated samples $G(\mathcal{Z})$ which leads to a better optimization behavior.

[1]Max Planck Institute for Intelligent Systems, Tübingen, Germany [2]Max Planck ETH Center for Learning Systems, Zürich, Switzerland. Correspondence to: Mehdi S. M. Sajjadi <msajjadi.com>.

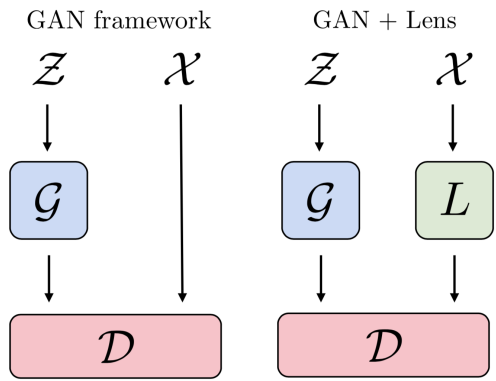

GAN framework     GAN + Lens

*Figure 1.* Schematic of the proposed module. We add a lens $L$ in between the real data $\mathcal{X}$ and the discriminator $D$. The lens is compatible with any type of GAN and dataset type. It finds a balance between fooling the discriminator and a reconstruction loss, leading to a tempered training procedure that self-adjusts to the capabilities of the current generator *w.r.t.* the current discriminator.

## 2. Tempered Adversarial Networks

The original formulation for GANs poses the training process as a minimax game between the generator $G$ and discriminator $D$ over the value function $\mathcal{V}$:

$$\min_G \max_D \mathcal{V}(D, G) = \mathbb{E}_{x \sim \mathcal{X}}[\log(D(x))]$$
$$+ \mathbb{E}_{z \sim \mathcal{Z}}[1 - \log(D(G(z)))] \quad (1)$$

In practice, both generator and discriminator are implemented as neural networks. The generator maps a random distribution $\mathcal{Z}$ to $G(\mathcal{Z})$ which is in the same space as the real data distribution $\mathcal{X}$. While the discriminator sees both real samples from $\mathcal{X}$ and fake samples from $G(\mathcal{Z})$, the generator only has control over the samples it produces itself, *i.e.*, it has no control over the real data distribution $\mathcal{X}$ which is fixed throughout training. To resolve this asymmetry, we add a lens module $L$ which modifies the real data distribution $\mathcal{X}$ before it is passed to the discriminator.

In practice, we use a neural network for $L$. The only change in the GAN architecture is consequently the input to the discriminator, which changes from $\{\mathcal{X}, G(\mathcal{X})\}$ to $\{L(\mathcal{X}), G(\mathcal{X})\}$.

We train the lens with two loss terms: an adversarial loss $\mathcal{L}_L^A$ and a reconstruction loss $\mathcal{L}_L^R$. The adversarial loss is supposed to maximize the loss of the discriminator of the respective GAN architecture, *i.e.*, $\mathcal{L}_L^A \approx -\mathcal{L}_D$. For the specific loss functions we used with the different GAN variants, see Sec. 2.1–2.3.

Additionally, we add a reconstruction loss to prevent the lens from converging to trivial solutions (*e.g.*, mapping all samples to zero):

$$\mathcal{L}_L^R = ||\mathcal{X} - L(\mathcal{X})||_2^2 \quad (2)$$

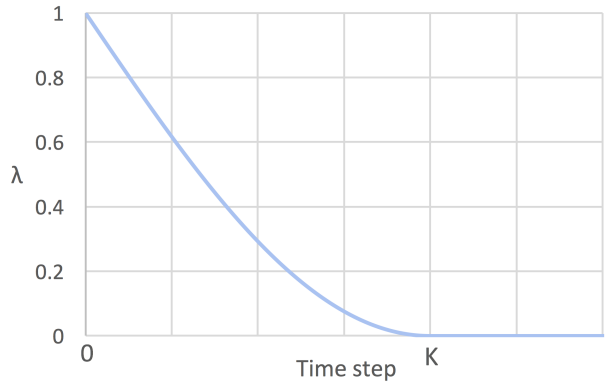

*Figure 2.* Schedule for the weight $\lambda$ for the adversarial loss term $\mathcal{L}_L^A$ of the lens during training. As training progresses, the value is lowered in a smooth way from 1 to 0 in $K$ steps, increasing the relative weight of the reconstruction loss for the lens. We set $K=10k$ in all experiments. While lower values showed faster convergence rates in our experiments, we opted for a single value in all experiments for simplicity and to avoid adding yet another hyperparameter that needs to be tuned. We found that the performance is robust against changed for the specific value for $K$ and a single value to yield good results across datasets and GAN architectures.

The overall loss for the lens is

$$\mathcal{L}_L = \lambda \mathcal{L}_L^A + \mathcal{L}_L^R \quad (3)$$

The lens can automatically balance a good reconstruction of the original samples with the objective of mapping the real data distribution $\mathcal{X}$ close to the generated data distribution $G(\mathcal{Z})$ *w.r.t.* the probabilities given by the discriminator. As training progresses, the generated samples get closer to the real samples, *i.e.*, the lens can afford to reconstruct the real data samples better. Once the discriminator starts to see differences, the loss term $\mathcal{L}_L^A$ increases which makes $L$ shift the real data distribution $\mathcal{X}$ towards the generated samples, helping to keep $G(\mathcal{Z})$ and $L(\mathcal{X})$ closer together which yields better gradients during training.

To accelerate this procedure, we set $\lambda = 1$ at the beginning of the training procedure and then gradually decrease it to $\lambda = 0$, at which point $L$ is only trained with $\mathcal{L}_L^R$, forcing it to converge to the identity mapping $L(\mathcal{X}) = \mathcal{X}$. To have a smooth transition from adversarial samples $L(\mathcal{X})$ to the real data distribution $\mathcal{X}$ in $K$ steps, we adapt the value for $\lambda$ as

$$\lambda = \begin{cases} 1 - \sin(t\pi/2K), & t \leq K \\ 0, & t > K \end{cases} \quad (4)$$

for the $t$-th time step during training. The value of $\lambda$ over time can be seen in Fig. 2. Once the lens converges to the identity mapping, training reduces to the original GAN architecture without a lens. In all experiments, we set $K = 10^5$ unless specified otherwise. Lower values for $K$ lead to faster convergence, but to avoid introducing a new hyperparameter that needs to be tuned, and for simplicity, we choose the same value for all experiments. Note that this choice is clearly not optimal for all tasks and tuning the value can easily lead to even faster convergence and higher quality samples.

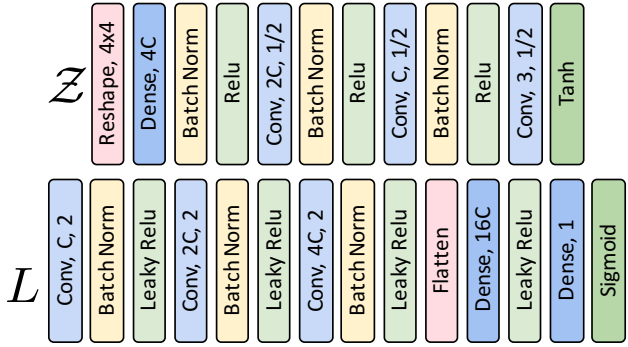

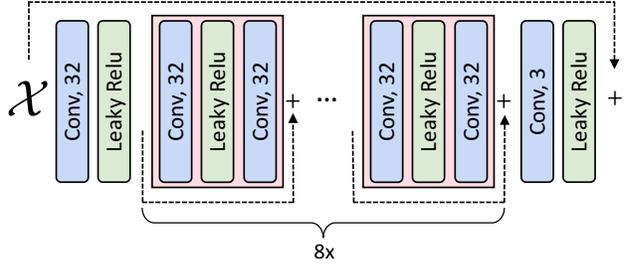

*Figure 3.* Network architecture of the generator $G$ (top) and discriminator $D$ (bottom). The design follows Radford et al. (2016). The strides of the convolutions are 1/2 for upsampling in $G$ and 2 for downsampling in $D$. The kernel size is 4×4 in both networks. The number of parameters can be varied by adjusting $C$.

### 2.1. Objectives for classical GAN formulation

In the original work, Goodfellow et al. (2014) use the loss

$$\mathcal{L}_G = -\log(D(G(\mathcal{Z}))) \qquad (5)$$

for the generator, and

$$\mathcal{L}_D^{original} = -\log(D(\mathcal{X})) - \log(1 - D(G(\mathcal{Z}))) \qquad (6)$$

for the discriminator. The objectives of generator and discriminator remain unchanged, though the input of the real data to $D$ is changed from $\mathcal{X}$ to $L(\mathcal{X})$:

$$\mathcal{L}_D = -\log(D(L(\mathcal{X}))) - \log(1 - D(G(\mathcal{Z}))). \qquad (7)$$

The lens is trained against the discriminator with the adversarial loss term

$$\mathcal{L}_L^A = -\log(1 - D(L(\mathcal{X}))) \qquad (8)$$

which minimizes the output of the discriminator for the lensed data points using the nonsaturating loss.

### 2.2. Objectives for LSGAN

In LSGAN (Mao et al., 2017), the log-loss is replaced by the squared distance. This leads to the adversarial loss

$$\mathcal{L}_G = ||D(G(\mathcal{Z})) - 1||_2^2 \qquad (9)$$

for the generator, and

$$\mathcal{L}_D = ||D(G(\mathcal{Z}))||_2^2 + ||D(L(\mathcal{X})) - 1||_2^2 \qquad (10)$$

*Figure 4.* Network architecture of the proposed lens that is similar to Sajjadi et al. (2017). The core of the network is composed of 8 residual blocks. To help convergence to identity, we add an additional residual connection from the input to the output. All convolutions have 3×3 kernels and stride 1.

for the discriminator. The lens works against the discriminator with the adversarial loss

$$\mathcal{L}_L^A = ||D(L(\mathcal{X}))||_2^2 \qquad (11)$$

### 2.3. Objectives for WGAN-GP

The discriminator or *critic* in the WGAN-GP variant (Gulrajani et al., 2017) outputs values that are unbounded, *i.e.*, there is no sigmoid activation at the after the last dense layer in Fig. 3. The objectives are

$$\mathcal{L}_G = -D(G(\mathcal{Z})) \qquad (12)$$

for the generator, and

$$\mathcal{L}_D = D(G(\mathcal{Z})) - D(L(\mathcal{X})) \qquad (13)$$

for the critic. Again, the lens works against the critic, so we use the adversarial objective

$$\mathcal{L}_L^A = D(L(\mathcal{X})) \qquad (14)$$

for the lens for this GAN variant.

### 2.4. Architecture, Training and Evaluation metrics

The lens can be any function which maps from the usually high-dimensional space of the real data distribution $\mathcal{X}$ to itself. Note that the lens does not need to be injective – in fact, early on during training, mapping several different points to the same data point can be a simple way to decrease the complexity of the data distribution which will likely decrease the loss term $\mathcal{L}_L^A$. Since it is desirable for the lens to turn into the identity mapping at some point during training, we have chosen a residual fully convolutional neural network architecture for the lens, see Fig. 4.

The network architecture and training procedure for the generator and discriminator depend on the chosen GAN framework. For the experiments with the original GAN loss, we use the DCGAN architecture along with its common tweaks (Radford et al., 2016), namely, strided convolutions instead of pooling layers, applying batch normalization in both networks, using ReLU in the generator and leaky ReLU in the discriminator, and Adam (Kingma & Ba, 2015) as the optimizer. See Fig. 3 for an overview of the networks. LSGAN is trained in the same setting but without batchnorm. For the WGAN-GP experiments, we used the implementation from Gulrajani (2017) which uses very similar models but the RMSProp optimizer (Hinton et al., 2012). We train the lens alongside the generator and discriminator and update it once per iteration regardless of the GAN variant. Note that the networks for the DCGAN and LSGAN experiments have intentionally been chosen not to have a very large number of feature channels to avoid memorization on small datasets which is why the results on an absolute scale are certainly not state of the art. We train using batch sizes of 32 and 64, a learning rate of $10^{-4}$ and we initialize the networks with the Xavier initialization (Glorot & Bengio, 2010).

For quantitative evaluation, previous works have been reporting the Inception score (Salimans et al., 2016), though its accuracy has been questioned (Barratt & Sharma, 2018). Recently, the *Fréchet Inception Distance* (FID) has been shown to correlate well with the perceived quality of samples, so we follow Heusel et al. (2017) and report FID scores. Note that a lower FID is better. For computational reasons, the FID scores are computed on sets of 4096 samples for the DCGAN and LSGAN experiments. While this is lower than the recommended 10k and should therefore not be compared directly with other publications, we found the sample size to be sufficient to capture relative improvements as long as sample sizes are identical. For the WGAN-GP experiments, we used sample sizes of 10k data points. The image size in all experiments is 32×32 pixels with 1 color channel for MNIST and 3 color channels for all other experiments.

## 3. Related works

After its introduction (Goodfellow et al., 2014), GANs have received a lot of attention from the community. There are several lines of work to improve the training procedure of GANs. Radford et al. (2016) proposed heuristic guidelines for the design of GAN architectures, *e.g.*, recommending the use of strided convolutions and batch normalization (Ioffe & Szegedy, 2015) in both generator and discriminator. Several works follow this trend, *e.g.*, Salimans et al. (2016) propose the use of further methods to stabilize the performance of GANs including feature matching, historical averaging, minibatch discrimination and one-sided label smoothing (Szegedy et al., 2016). More closely related to our work, Arjovsky & Bottou (2017) and Mehrjou et al. (2017) propose adding noise to either both real and fake or only to the real samples during training with the motivation of increasing the support of the generated and real data distributions which leads to more meaningful gradients. The amount of noise is reduced manually during training. In our work, the lens is not constrained in the mapping that it can apply to balance the training procedure. Furthermore, the effect of the lens is automatically balanced with a reconstruction term that adjusts the intervention of the lens dynamically during training depending on the current balance between generator and discriminator.

There are several works which approach the problem by using multiple networks instead of one. Denton et al. (2015) propose a Laplacian pyramid of generator-discriminator pairs for generating images. Zhang et al. (2017) use a similar approach by using one GAN to produce a low-resolution image and another GAN which produces higher-resolution images conditioned on the output of the low-resolution GAN. Such methods have the drawback that several GANs need to be trained which increases the number of parameters and introduces a computational bottleneck. Most recently, Karras et al. (2018) produced convincing high-resolution images of faces by first learning the low frequencies in images and then progressively growing both networks to produce higher-resolution images. While promising, all of the methods above are constrained to generating images since the concept of resolution is not easily generalizable to other domains.

Another line of research attacks the problem of training GANs by changing the loss functions, *e.g.*, Mao et al. (2017) use the least-squares distance loss whereas Arjovsky et al. (2017) approximate the Wasserstein distance which provides more stable gradients for the generator. Gulrajani et al. (2017) improve upon the latter by replacing weight clipping in the discriminator with a gradient penalty which accelerates the training procedure considerably.

In the context of training neural networks, Gulcehre et al. (2017) smoothen the objective function by adding noise to activation functions and then gradually decrease the level of noise as training progresses. Bengio et al. (2009) coin the term *curriculum learning* where the idea is to present the samples during training in a specific order that improves the learning process. Our approach may have a similar effect, but differs in that we present all samples of the original dataset to the networks, modifying them dynamically in a way that stabilizes the learning process.

## 4. Experiments

Showing that modifications or additions to GANs lead to *better* results in any way is a delicate topic that has raised much controversy in the community. Most recently, the findings of Lucic et al. (2017) suggest that with a sufficient computational budget, any GAN architecture can be shown to perform at least as well or better than another, if a smaller computational budget is spent on the hyperpa-rameter search for the latter. To avoid this fallacy and to prevent choices such as the network architecture or chosen hyperparameters to favor one or another method, we follow common guidelines that are currently in use for training GANs and we conduct experiments with three different GAN frameworks: the original GAN formulation by Goodfellow et al. (2014); LSGAN, where Mao et al. (2017) replace the log-loss with the least-squares loss; and WGAN-GP, where Gulrajani et al. (2017) minimize the approximated Wasserstein distance between real and generated data distributions and where the training procedure includes a gradient penalty for the discriminator. For the network architecture, we follow standard design patterns (see Sec. 2.4). In our experimental section, we do not strive for state of the art in the end results, but rather we test how much of an effect the lens can have on training. We show that the simple addition of a lens can help improve results across various GAN frameworks. We hope that this insight will help ongoing efforts to understand and improve the training of GANs and other neural network architectures.

In all experiments, the random weights for the initialization of the networks were identical for the GANs with and without a lens. All experiments have further been run with at least 3 different random seeds for the weight initialization to prevent chance from affecting the results.

## 4.1. Original GAN objective

### 4.1.1. MNIST

We begin with the original GAN variant on the classical MNIST dataset. To analyze the behavior of the lens, we first consider the case of a fixed $\lambda = 1$, *i.e.*, the lens has no direct incentive to become perfect identity. Fig. 5 (top) shows generated and lensed samples at different training stages for this architecture. At the beginning of training, the lens scrambles the MNIST digits to look more similar to the generated images. As the generator catches up and produces digit-like samples, the lens can afford to improve reconstruction. Since the lens acts as a balancing factor between the $G$ and $D$, this leads to a very stable training procedure. However, even after 10M steps, the reconstruction of the lens still improves, as does the FID score of the generated samples (see FID plot in Fig. 5, bottom left). In comparison, the GAN without a lens converges much faster to better FID scores (Fig. 5, bottom right, green curve).

To accelerate the training procedure, we adapt the weight of $\lambda$ as explained in Sec. 2. As this forces the lens to turn into a perfect identity mapping at some point during training, the process converges much more quickly and easily surpasses the quality of the GAN without a lens, yielding FID scores of 22 (with lens) vs. 42 (without lens). Additional experiments with much larger, heavily fine-tuned architectures that already show stable training for GANs did not show better FID after the addition of the lens, indicating that the proposed method can stabilize weaker architectures and lead to more robust GAN training with respect to hyperparameters.

### 4.1.2. Color MNIST

Since MNIST only has 10 main modes, it is not an adequate test for the mode collapse problem in GANs. To alleviate this, a color MNIST variant has been proposed (Srivastava et al., 2017). Each sample is created by stacking three randomly drawn MNIST digits into the red, green and blue channels of an RGB image which leads to a dataset with 1000 modes (assuming 10 modes for MNIST) while still being easy to analyze visually.

As can be seen in Fig. 9, the GAN without a lens first produces decent results in all color channels before it collapses partially. At this point, only the green color channel looks like MNIST digits while the other two channels are clearly not from the correct distribution. The FID reflects this, sometimes even increasing as training proceeds, with values throughout training never getting lower than 50. Adding the lens to the GAN stabilizes training and leads to much higher quality samples with an FID of 9 for the best samples compared to 53 for the GAN without a lens.

## 4.2. LSGAN objective

### 4.2.1. MNIST

We found the LSGAN variant to be sensitive to the random seed for the weight initialization of the networks. LSGAN without a lens did not train in most cases, with the best run yielding FID scores of 19. With the lens, the networks always trained well, with the worst run producing FID scores of 16 and the best run giving FID scores of 14.

### 4.2.2. Color MNIST

On the Color MNIST dataset, we found LSGAN to perform similarly. The best run without a lens yielded FID scores of 90 and training stalled there due to starved gradients. Adding the lens made the networks produce meaningful results in all runs, producing FID scores between 14 and 22 from different random initializations.

### 4.2.3. Celeba

On the CelebA dataset (Liu et al., 2015), LSGAN was unstable, with a starving generator early on during training due to a perfect discriminator that did not provide gradients. The best run without a lens yielded an FID score of 52. Adding the lens helped the system stabilize and produce meaningful results in all runs, with the best run yielding FID scores of 32 and the worst run yielding an FID of 37. Note that these numbers are comparably high due to the small model size of the generator and discriminator. The effects of the lens during training are shown in Fig. 10.

Generated samples $G(\mathcal{Z})$            Lensed samples $L(\mathcal{X})$

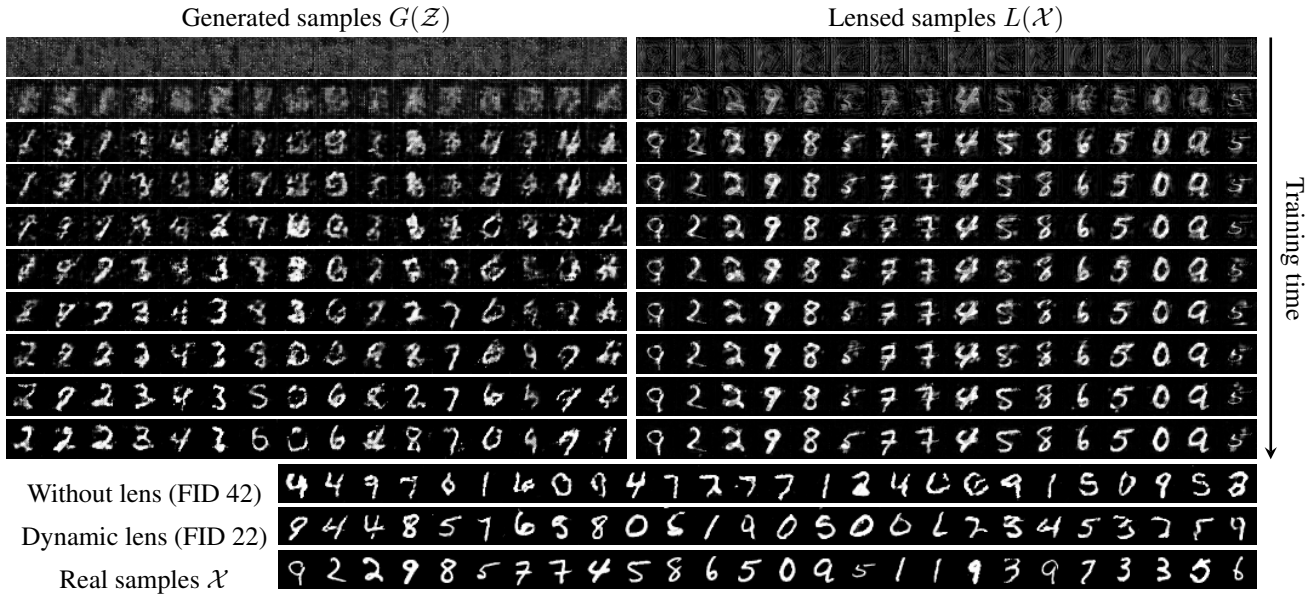

Without lens (FID 42)

Dynamic lens (FID 22)

Real samples $\mathcal{X}$

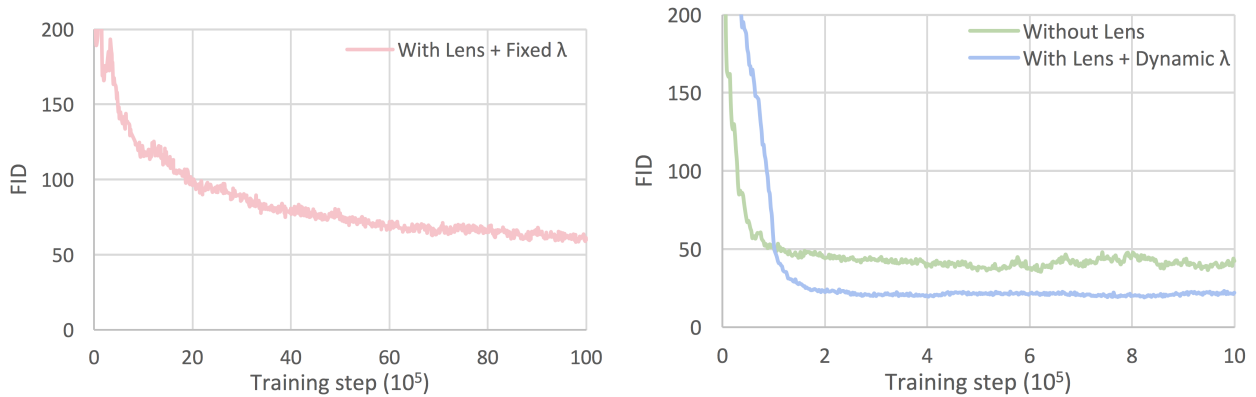

*Figure 5.* MNIST digits produced by DCGAN with a lens with **fixed** $\lambda = 1$ (top). The columns show generated and lensed samples. The lens $L$ adds perturbations that make the real data samples look more similar to fake samples. As training progresses, the quality increases and the reconstruction of $L$ improves steadily. Ideally, the system would converge to a point where $G$ produces samples that are indistinguishable from $\mathcal{X}$ for a fully trained discriminator – at this point, $L$ would turn into the identity mapping. While training with the lens is very stable and while the FID was still decreasing when we stopped training, the reconstructions are not perfect even after 10M training steps and the FID is still only 60, *i.e.*, it has not yet even reached the performance of DCGAN without a lens after only 1M steps (bottom right, green curve). When the value for $\lambda$ is adapted (see Sec. 2), training is greatly sped up and, the quality of the samples is substantially higher (FID 22) than for the GAN without a lens (FID 42). The difference is also visible in the results, where the GAN with a lens produces better looking MNIST digits. Note that the FID is initially higher for the GAN with a lens in the bottom right. This is because the FID is always measured against the real samples $\mathcal{X}$, while $G$ is initially trained for the lensed distribution $L(\mathcal{X})$ that differs from $\mathcal{X}$ in the early training stages.

## 4.3. WGAN-GP

### 4.3.1. CIFAR10

To test the lens on an entirely different GAN architecture, we also add it to the WGAN-GP framework (Gulrajani et al., 2017). Wasserstein GANs are generally believed to be more stable than other GAN variants, making it harder for tweaks to significantly improve sample quality. Never-theless, our experiment on the Cifar-10 dataset shows that the same lens with the same hyperparameters also works well with WGAN-GP, yielding higher-quality results as measured by the FID score at an earlier training stage. As seen in Fig. 7, the model with a lens quickly surpasses the quality of the model without a lens and it takes some more training time for the GAN without a lens to catch up. When trained long enough, both models yield an FID of 39.

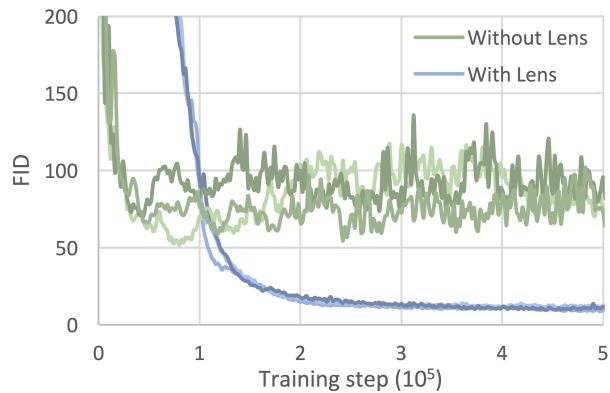

Figure 6. FID for DCGAN trained on the Color MNIST dataset. For each method, 3 independent runs with different random seeds for the weight initialization are shown. Since the value of $\lambda$ is high early on during training, the GAN with a lens initially performs worse, but the quality soon catches up and surpasses that of the GAN without a lens as $\lambda$ is lowered to a value of 0. The GANs with a lens are much more stable and more robust against different random seeds for the weight initialization.

It is noteworthy that adding the lens can lead to faster training although the generator and discriminator are initially trained on a data distribution $L(\mathcal{X})$ that is quite different from the real data distribution $\mathcal{X}$ (see Fig. 8). This result suggests that a scheduled learning procedure can indeed accelerate optimization of neural networks. The proposed lens is a natural way to dynamically adjust the rate at which learning proceeds.

## 5. Conclusion

We propose a generic module that leads to a dynamically self-adjusting progressive learning procedure of the target data distribution in GANs. A number of experiments on several GAN variants highlight the potential of this approach. Whilst the method is conceptually simple, it may have significant potential, not only in the image domain, but also in other domains such as audio or video generation. We hypothesize that similar modifications can be applied to improve optimization of other neural network architectures. For instance, autoencoders can be tempered by initially training to reconstruct lensed inputs, and recognition networks can be tempered by grouping or smoothing classes. Finally, it may be possible to incorporate prior knowledge about the task at hand by suitably biasing or initializing lenses, for instance using blurring lenses to generate images starting from low-frequency approximations.

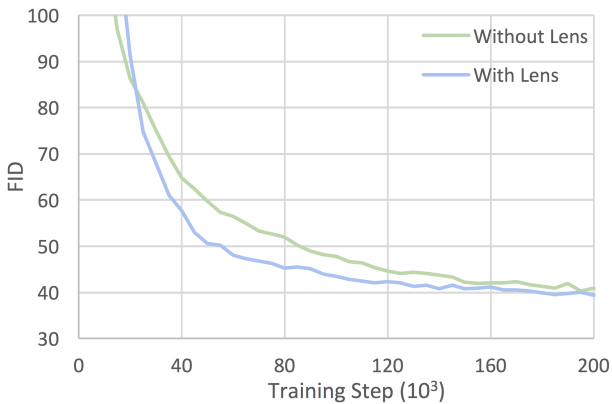

Figure 7. FID for WGAN-GP on Cifar-10 with and without a lens. The value for $\lambda$ is smoothly lowered from 1 to 0 in the first $K$=10K steps. The final results have similar FIDs, but WGAN-GP with a lens converges faster to higher-quality samples. Tuning the rate at which $\lambda$ is adapted could further improve convergence speeds.

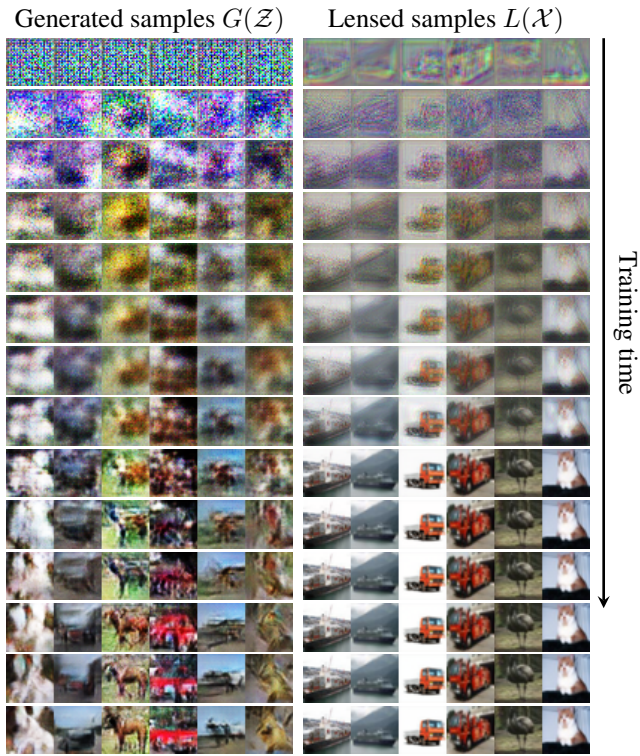

Figure 8. WGAN-GP with a lens $L$. In early training stages, the images are blurry lack contrast, but $L$ gradually reconstructs finer details as $G$ catches up. Note that by design, $L$ could easily converge to the perfect identity mapping very quickly, so the gradual improvements seen here are a result of the adversarial loss term $\mathcal{L}_L^A$ rather than slow convergence.

Generated samples $G(\mathcal{Z})$               Lensed samples $L(\mathcal{X})$

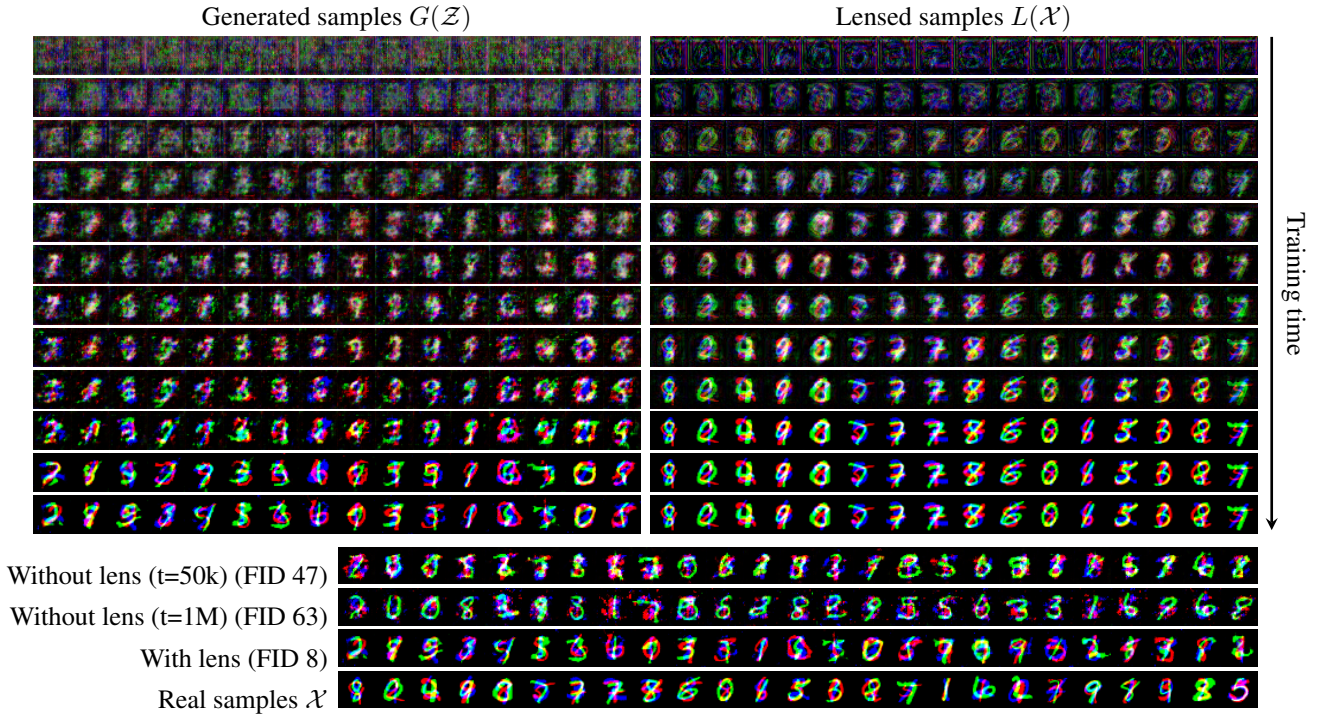

Without lens (t=50k) (FID 47)
Without lens (t=1M) (FID 63)
With lens (FID 8)
Real samples $\mathcal{X}$

*Figure 9.* Results of DCGAN with a lens $L$ on the Color MNIST dataset (top). The lens gradually improves reconstruction as $G$ produces better samples. Once $L$ is a perfect identity function, $G$ adds remaining details and finally produces realistic results (bottom, third row). In comparison, the GAN without a lens only manages to produce good-looking digits in the green color channel and produces noise in the red and blue channels (bottom, first row, t=50k). As $G$ improves quality in the green channel, the quality in the other two channels decreases (bottom, second row, t=1M) which is a commonly encountered instability during GAN training. Several runs with different random seeds for the weight initialization yielded similar results for both architectures, see Fig. 6. Images best viewed in color.

Generated samples $G(\mathcal{Z})$               lensed samples $L(\mathcal{X})$

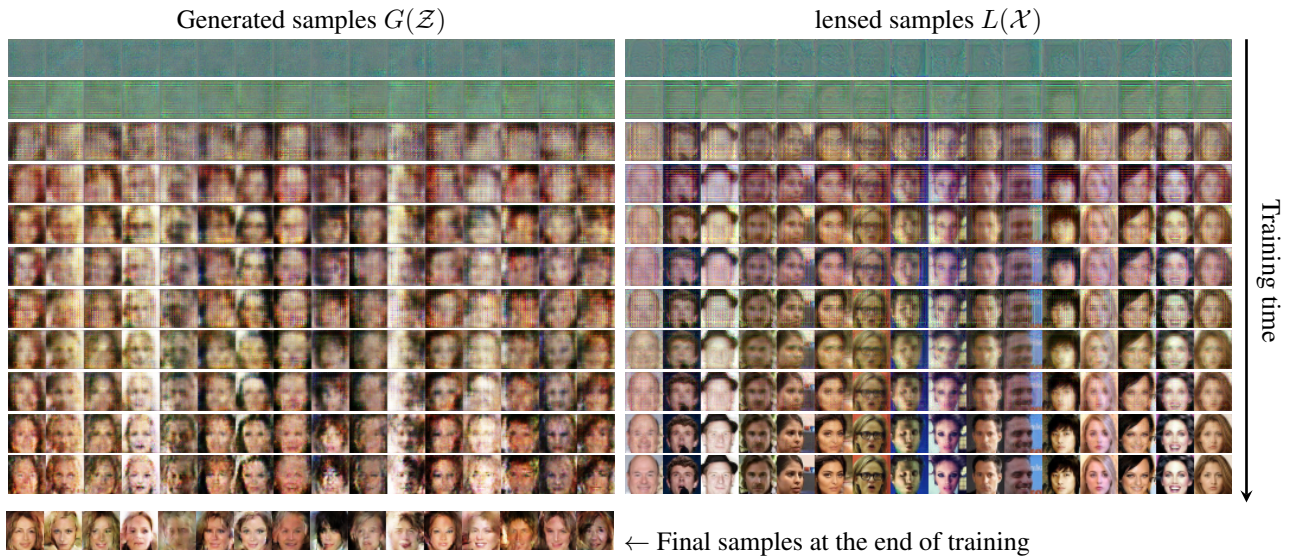

← Final samples at the end of training

*Figure 10.* Generated and lensed samples at various steps during the training process of LSGAN on the CelebA dataset with a lens. The generator produces a large variety of faces since it is not forced to reproduce fine details early during training, making it less prone to the mode collapse problem.